\documentclass[a4paper,twoside]{article}

\usepackage{epsfig}
\usepackage{subfigure}
\usepackage{calc}
\usepackage{amssymb}
\usepackage{amstext}
\usepackage{amsmath}
\usepackage{amsthm}
\usepackage{multicol}
\usepackage{pslatex}
\usepackage{apalike}
\usepackage{hyperref}
\usepackage{SCITEPRESS}     

\newenvironment{bulletlist}
  {\begin{list}{$\bullet$}
     {\setlength{\parsep}{6pt}
      \setlength{\topsep}{6pt}
      \setlength{\itemsep}{6pt}}}
  {\end{list}}
\begin{document}

\title{A continuum among logarithmic, linear, and exponential functions,
	and its potential to improve generalization in neural networks}

\author{\authorname{Luke B. Godfrey and Michael S. Gashler}
\affiliation{Department of Computer Science and Computer Engineering}
\affiliation{University of Arkansas, Fayetteville, AR, USA}
\email{\{lbg002, mgashler\}@uark.edu}
}


\abstract{
We present the soft exponential activation function for artificial neural networks that continuously interpolates between logarithmic, linear, and exponential functions.
This activation function is simple, differentiable, and parameterized so that it can be trained as the rest of the network is trained.
We hypothesize that soft exponential has the potential to improve neural network learning, as it can exactly calculate many natural operations that typical neural networks can only approximate, including addition, multiplication, inner product, distance, polynomials, and sinusoids.
}

\onecolumn \maketitle \normalsize \vfill

\section{\uppercase{Introduction}}\label{sec_introduction}

Each neuron in an artificial neural network applies a non-linear activation function to a weighted sum of its inputs.
The activation function serves the important role of enabling the neural network to fit to non-linear curves and surfaces.
If omitted, even deep multi-layered neural networks reduce to be functionally equivalent to simple linear regression.
Hence, the activation function endows the neural network with its representational power.

One might ask, \emph{which activation function is best for neural networks?}
For years, the logistic and tanh functions have been popular choices \cite{kalman1992tanh}.
More recently, rectified linear units have been shown to possess desirable properties \cite{nair2010rectified,zeiler2013rectified}.
While these functions perform well empirically, little theoretical basis has been found to justify their extensive use over many other potential functions.
We present the soft exponential function, a novel activation function with many desirable theoretical properties.
It continuously interpolates between logarithmic, linear, and exponential activation functions.
It enables neural networks to exactly compute many natural mathematical structures that can only be approximated by neural networks that use traditional activation functions,
including addition, multiplication, exponentiation, dot product, Euclidean and L-norm distance, polynomials, Gaussian radial basis functions, and Fourier neural networks.

The next section derives soft exponential and the remainder of the paper discusses its desirable properties.

\section{\uppercase{Derivation}}\label{sec_solution}

It is well known that multiplication can be implemented by means of addition in logarithmic space.
That is,

\begin{equation}
	p*q = e^{(\log_e p) + (\log_e q)}.
	\label{eq_logadd}
\end{equation}

This property can enable neural networks that use a mixture of logarithmic, linear, and exponential activation functions to exactly perform the basic mathematical operation of multiplication.
However, using a mixture of different activation functions in a single neural network adds a significant component of complexity.
Specifically, it leaves the user to determine which activation function should be used with each neuron in the network.
If a function can be found that continuously generalizes between logarithmic, linear, and exponential functions, then a neural network with a single activation function would be empowered to autonomously learn to add, multiply, exponentiate, and compute the logarithms as needed to accomplish arbitrary tasks.
Because these mathematical operations have proven to have significant value in nearly all other areas of science, it is natural to suppose that neural networks should be given the ability to perform the same operations when they attempt to autonomously model various phenomena.

A simple equation that continuously interpolates between
linear and exponential functions is

\begin{equation}
	g(\alpha,x) = \frac{e^{\alpha x} - 1}{\alpha} + \alpha.
	\label{eq_exp}
\end{equation}

Note that $\lim_{\alpha \to 0}g(\alpha,x)=x$, and $g(1,x)=e^x$.
This function does not become a logarithmic function (i.e. when $\alpha=-1$), so it does not provide a complete solution to our objective.
However, we can invert $g$ with respect to $x$ to obtain a function that interpolates between logarithmic and linear functions:

\begin{equation}
	g^{-1}(\alpha,x) = \frac{\log_e(1+\alpha (x - \alpha))}{\alpha}.
	\label{eq_log}
\end{equation}

Since $g$ and $g^{-1}$ are equivalent when $\alpha=0$, we can mathematically piece
them together along that edge without breaking continuity.
We negate $\alpha$ in the case of the inverse function and obtain the following continuous piecewise function:

\begin{equation}
	f(\alpha,x) = \left \{
		\begin{array}{rcl}
			-\frac{\log_e(1-\alpha (x + \alpha))}{\alpha} & \mbox{for} & \alpha < 0 \vspace{1mm} \\
			x & \mbox{for} & \alpha = 0 \vspace{1mm} \\
			\frac{e^{\alpha x} - 1}{\alpha} + \alpha & \mbox{for} & \alpha > 0.
		\end{array} \right.\\
	\label{eq_logexp}
\end{equation}

Equation~\ref{eq_logexp} interpolates between logarithmic, linear, and exponential functions.
Although it is spliced together, it is continuous both with respect to $\alpha$ and with respect to $x$, and has a number of properties that render it particularly useful as a neural network activation function.
We call $f$ the soft exponential activation function.

We can now address the challenge of creating a continuum of operations between addition and multiplication.
By substituting $f$ into Equation~\ref{eq_logadd}, we obtain a continuous generalization between these two operations:

\begin{equation}
	h(\beta, p, q)=f\left(\beta, f(-\beta, p) + f(-\beta, q) \right).
\end{equation}

\begin{figure}[!tb]
	\begin{center}
		\includegraphics[width=2.8in]{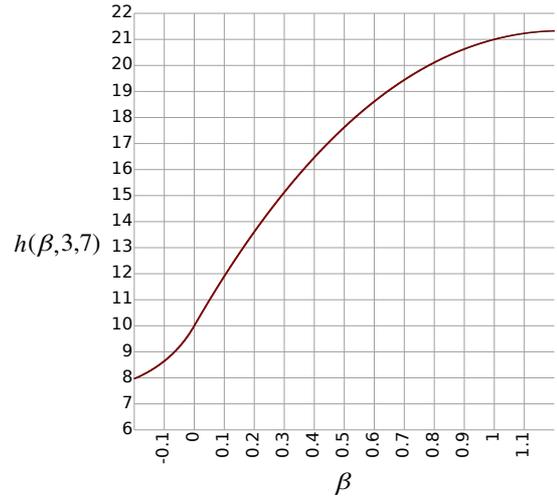}
		\caption{A plot of $h(\beta, 3, 7)$.
			When $\beta=0$, it correctly calculates $3+7=10$.
			When $\beta=1$, it correctly calculates $3*7=21$.}
		\label{fig_interp}
	\end{center}
\end{figure}

If $\beta=0$, this function adds $p$ and $q$. If $\beta=1$, it multiplies $p$ and $q$.
Figure~\ref{fig_interp} illustrates this continuum between addition and multiplication
with the arbitrary values $p=3$ and $q=7$.
At $\beta=0$, it correctly calculates $3+7=10$, and at $\beta=1$, it correctly calculates $3*7=21$.

\section{\uppercase{Analysis}}\label{sec_analysis}

\begin{figure}[!tb]
	\begin{center}
		\includegraphics[width=2.8in]{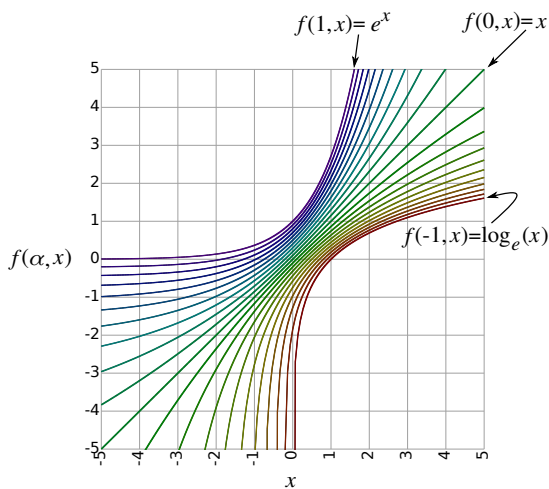}
		\caption{A plot of $f(\alpha, x)$ for $\alpha=\{-1,-0.9,-0.8,\cdots,0.8,0.9,1.0\}$ from red to purple.}
		\label{fig_activation}
	\end{center}
\end{figure}

\begin{figure}[!tb]
	\begin{center}
		\includegraphics[width=2.8in]{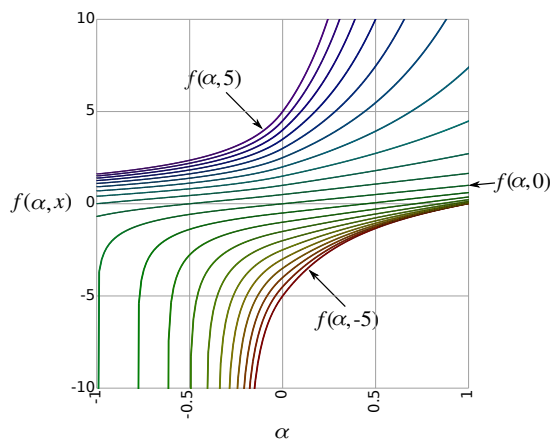}
		\caption{A plot of $f(\alpha, x)$ for $x=\{-5,-4.5,-4,\cdots,4,4.5,5\}$ from red to purple.}
		\label{fig_activation2}
	\end{center}
\end{figure}

Some of the nice properties of soft exponential include:

\begin{bulletlist}
	\item $f(-1,x)=\log_e(x)$
	\item $f(0,x)=x$
	\item $f(1,x)=e^x$
	\item For other values of $\alpha$, $f(\alpha,x)$ does something continuous and reasonable.
	\item The equation is simple, and can be implemented in code with very few operations.
	\item It appears reasonably smooth when plotted. (See Figures~\ref{fig_activation} and \ref{fig_activation2}.)
	\item Negating $\alpha$ inverts the function, such that $f^{-1}(\alpha,x)=f(-\alpha,x)$.
	\item For any constant value of $\alpha$, $f(\alpha,x)$ is monotonic.
	\item It is continuously differentiable with respect to $x$,\\
\begin{equation}
	\frac{\partial f}{\partial x} = \left \{
		\begin{array}{rcl}
			\frac{1}{1-\alpha (\alpha + x)} & \mbox{for} & \alpha < 0  \vspace{1mm}\\
			e^{\alpha x} & \mbox{for} & \alpha \ge 0
		\end{array} \right.\\
\end{equation}
		because
$$
		\lim_{\alpha \to 0^+}\frac{\partial f}{\partial x}\equiv \lim_{\alpha \to 0^-}\frac{\partial f}{\partial x} \equiv 1.
$$
	\item And it is continuously differentiable with respect to $\alpha$,\\
\begin{equation}
	\frac{\partial f}{\partial \alpha} = \left \{
		\begin{array}{rcl}
			\frac{\log_e(1-(\alpha^2+\alpha x))-\frac{2\alpha^2+\alpha x}{\alpha^2+\alpha x-1}}{\alpha^2} & \mbox{for} & \alpha < 0 \vspace{1mm}\\
			\frac{x^2}{2}+1 & \mbox{for} & \alpha = 0 \vspace{1mm}\\
			\frac{\alpha^2 + (\alpha x - 1) e^{\alpha x} + 1}{\alpha^2} & \mbox{for} & \alpha > 0
		\end{array} \right.\\
\end{equation}
		because
$$
		\lim_{\alpha \to 0^+}\frac{\partial f}{\partial \alpha} \equiv \lim_{\alpha \to 0^-}\frac{\partial f}{\partial \alpha} \equiv \frac{x^2}{2}+1.
$$
	\item Because it is differentiable, it is possible to train a neural network with soft exponential using gradient descent. The alpha parameter of the activation function is updated in the same manner as the weights, by stepping in the gradient direction that reduces some objective function.
\end{bulletlist}

\section{\uppercase{Inner product}}\label{sec_inner_product}

\begin{figure}[!tb]
	\begin{center}
		\includegraphics[width=2.5in]{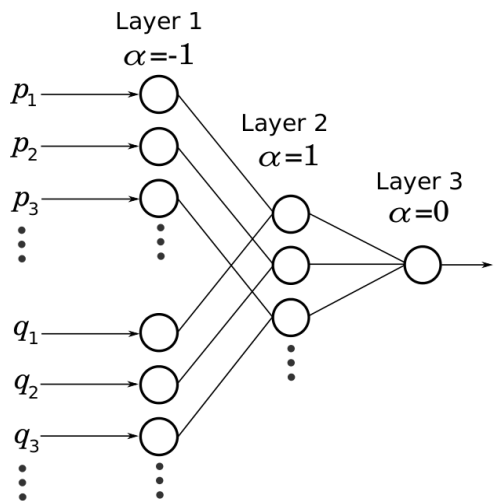}
		\caption{A neural network implementation of inner product using soft exponential as an activation function.
			All of the weights represented with lines in this figure have a value of 1.
			All other weights have a value of 0.}
		\label{fig_neuralnet}
	\end{center}
\end{figure}

Another operation we might want to generalize is inner product.
The inner product is typically implemented as,
$\mathbf{p}\cdot\mathbf{q}=p_{0}q_{0}+p_{1}q_{1}+p_{2}q_{2}+\ldots$.
Inner product could be implemented using a 3-layer neural network as depicted in Figure~\ref{fig_neuralnet}.
This network uses soft exponential for the activation function in each of its units.
The first layer computes the logarithm of all the elements in $\mathbf{p}$ and $\mathbf{q}$.
(All the units in this layer use $\alpha=-1$.)
The second layer adds corresponding elements of $\mathbf{p}$ and $\mathbf{q}$, and exponentiates the result.
(All the units in this layer use $\alpha=1$.)
The third layer sums all the pair-wise products together.
(The unit in this layer uses $\alpha=0$.)

One possible use for this generalization of inner product is to implement a neural network version of
matrix factorization, a useful algorithm for recommender systems \cite{koren:matrix_factorization} and missing value imputation for sparse matrix completion \cite{cai:matrix_completion}.
Matrix factorization has also proved to be effective for document clustering \cite{xu:document}, text mining and spectral data analysis \cite{berry:algorithms}, and molecular pattern discovery \cite{brunet:metagenes}.
A neural network with our activation function can exactly compute inner product and matrix factorization, and thus it should be able to achieve accuracy at least as good as approaches that do not use neural networks.
Because of the flexibility of this generalized approach, it has the potential to outperform direct matrix factorization.
For example, in a recommender system, our approach facilitates augmenting user and item profile vectors with static profile vectors for addressing the cold-start problem \cite{koren:matrix_factorization}.

\section{\uppercase{Distance}}\label{sec_distance}

Suppose we want to compute the distance between two vectors, $\mathbf{p}$ and $\mathbf{q}$.
This could also be done with a neural network that uses soft exponential for its activation functions.
To do this, we will use the property,
$$
	a^b = e^{b\log_e(a)}.
$$
Figure~\ref{fig_sqdistance} shows a neural network that computes the squared distance between two vectors.
(If you want to take the square root, to make it Euclidean distance,
just change the unit in layer 3 to use $\alpha=-1$, and add a layer 4 with one unit.
This unit would use $\alpha=1$, and its incoming weight would be set to 0.5.)

\section{\uppercase{Polynomials}}\label{sec_polynomials}

\begin{figure}[!tb]
	\begin{center}
		\includegraphics[width=2.5in]{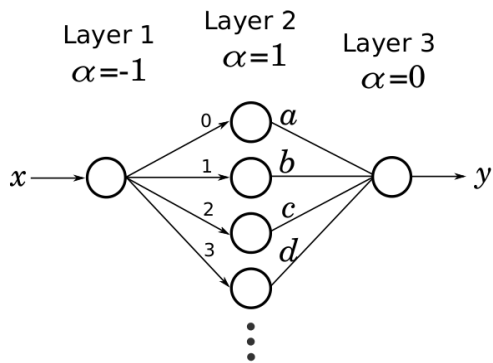}
		\caption{A neural network implementation of a polynomial, $y = a + bx + cx^2 + dx^3 \ldots$, using soft exponential as an activation function.}
		\label{fig_polynomial}
	\end{center}
\end{figure}

Figure~\ref{fig_polynomial} shows a neural network that exactly computes an arbitrary polynomial.
Multivariate polynomials could also be implemented by simply adding additional units on the input end.

\section{\uppercase{Radial basis function networks}}\label{sec_rbm}

\begin{figure}[!tb]
	\begin{center}
		\includegraphics[width=2.5in]{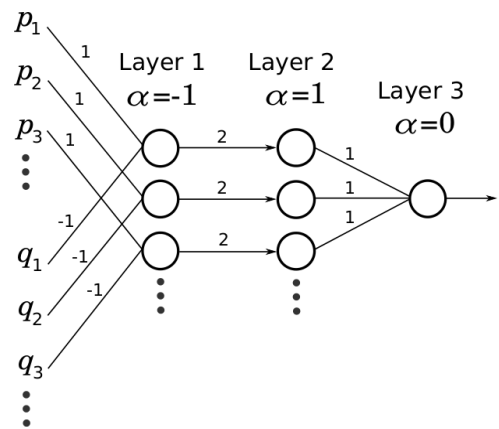}
		\caption{A neural network implementation of squared distance using soft exponential as an activation function.
			To compute Euclidean distance (the square root of this), only one additional network unit would be required.}
		\label{fig_sqdistance}
	\end{center}
\end{figure}

A gaussian radial basis kernel uses the formula,
$$
	e^{-rs},
$$
\noindent where $r$ is a weight that controls the squared radius of the kernel,
and $s$ is either the squared distance between the input vector and the center of the kernel,
or the inner product with the input vector.
This function is important to a number of classification models, including support vector machines that use a radial basis function and radial basis function networks \cite{scholkopf1997comparing,chen1991orthogonal,qasem2011radial}.
This could be implemented in a network using only $f$ as an activation function by simply
adding a single unit with $\alpha=1$ to the neural networks in Figures \ref{fig_neuralnet} or ~\ref{fig_sqdistance}.
The weight feeding into this unit would be $-r$.
If we added a layer to combine several of these,
we would have a radial basis function network without using any specialized units.

Although it is already well-known that neural networks are universal function approximators \cite{cybenko1989ann_universal_function_approximators}, it is worth noting that soft exponential enables common architectures to be exactly implemented using a neural network with minimal architectural overhead.
If a simple model sufficiently models a set of data, it is generally preferable and yields better predictions than an unnecessarily complex one.
If these architectures were implemented using a network with a sigmoidal activation function, for example, the resulting models would be very large networks that would probably take more training data to train it to generalize well.

\section{\uppercase{Fourier networks}}\label{sec_fourier}

Fourier neural networks use a sinusoidal activation function to transform a signal from the time or space domain to the frequency domain in a process similar to the Fourier transform \cite{silvescu1999fourier,tan2006fourier,zuo2009fourier}.
If $\alpha$ is allowed to have a complex value, soft exponential can be used as the activation function in a Fourier neural network.
Let $\alpha_r$ be the real component of $\alpha$, and $\alpha_i$ be the imaginary component of $\alpha$,
such that $\alpha=\alpha_r + i \alpha_i$.
For simplicity, we assume that $x$ is real, and $\alpha_r = 0$.
Then the equation for $f$ becomes

\begin{equation}
	f(\alpha_i,x) = \frac{\sin(\alpha_i x)}{\alpha_i} + i \left(\alpha_i - \frac{\cos(\alpha_i x)+1}{\alpha_i}\right).
	\label{eq_real}
\end{equation}

\begin{figure}[!tb]
	\begin{center}
		\includegraphics[width=2.8in]{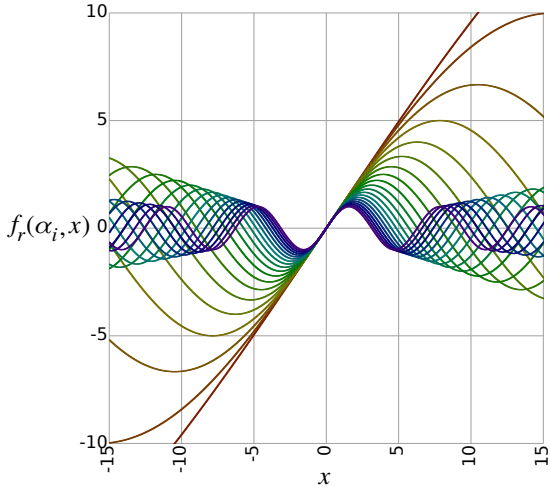}
		\caption{A plot of the real component of soft exponential over a range of values for $\alpha_i$.}
		\label{fig_fourier}
	\end{center}
\end{figure}

\begin{figure}[!tb]
	\begin{center}
		\includegraphics[width=2.8in]{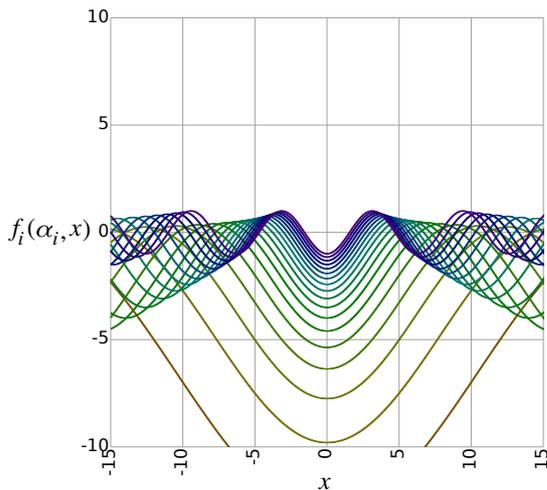}
		\caption{A plot of the imaginary component of soft exponential over a range of values for $\alpha_i$.}
		\label{fig_fourier2}
	\end{center}
\end{figure}

Without these assumptions, the resulting equation contains several additional terms.
Figures~\ref{fig_fourier} and \ref{fig_fourier2}
show the real and imaginary components respectively of $f$ over a range of values for $\alpha_i$.
It can be seen in these figures that the imaginary component of $\alpha$ determines the frequency of the sinusoidal wave.
(Although it also affects the amplitude, this is not significant because the outgoing weight can compensate to achieve any desired amplitude).

We have shown that Fourier networks are effective for extrapolating real-world time-series data \cite{gashler2014fourier}.
In a pending publication, we showed that this approach is even more effective at generalizing when it it combined with other activation functions \cite{godfrey2015neural}.
Because soft exponential can be logarithmic, exponential, linear, or sinusoidal when $\alpha$ is allowed to be complex, we can create a Fourier network with only this activation function and achieve the same level of accuracy for generalization and extrapolation.

\section{\uppercase{Proposed architecture}}\label{sec_architecture}

We conclude our discussion by describing a deep neural network architecture that
could potentially use this novel activation function to autonomously achieve all of these representational capabilities as needed to address a wide range of challenges.
Because complex values for $\alpha$ cause each unit to output two values, instead of one,
it may not be immediately clear how to apply such a network to arbitrary problems.
However, if the $\alpha$ parameter values in the output layer are constrained to take only real values,
then this network will behave like traditional neural networks, mapping from any number of input values to any number of output values.
Allowing hidden units to take on complex values for $\alpha$ should not present any problems because
the additional values may simply be fed into the next layer as if the preceding layer were twice as big.
Hence it should be reasonable to use $f$ as the activation function for every unit in a deep neural network.

The $\alpha$ parameter for each unit could be initialized to $0 + 0i$.
This has the very desirable property of initially causing the entire network to behave like linear regression.
As training proceeds, it will take on non-linearities only as necessary to fit the data.
All of the weights would be initialized with random values drawn from a normal distribution,
then normalized such that the primary eigenvalue is 1.
Since all of the activation functions are initially the identity function,
the problem of vanishing gradients is initially mitigated, enabling very deep networks to be trained efficiently.
This activation function does not impose any particular topology on the rest of the network, so 
the layers could fully-connected or arranged with sparse connections, such as in convolutional layers.

Likewise, the differentiability of soft exponential facilitates optimization with batch gradient descent, stochastic gradient descent, or many other optimization techniques.
$\alpha$ can be updated along with the weights in the manner of steepest descent.
$L^1$ regularization should be applied to promote sparsity.
It can be observed that the various common architectures that we can demonstrated with this activation function use sparse connections.
It follows, therefore, that $L^1$ regularization may be expected to work particularly well with this activation function.
Note that $L^1$ regularization can be applied to the $\alpha$ parameter as well as the weights of the network.
When $\alpha$ is pulled toward zero, the network approaches linear regression.
Hence, regularizing the $\alpha$ parameter has the desirable effect of causing the surface represented by the neural network to straighten out.

\section{\uppercase{Conclusion}}\label{sec_conclusion}

We presented a novel activation function, soft exponential, that continuously generalizes among logarithmic, linear, and exponential functions.
This function exhibits many desirable theoretical properties that make it well-suited for use as an activation function with neural networks.
Empirical validation of these theoretical properties still needs to be performed as future work.
Because of the significant potential that this activation function has to impact the effectiveness of deep neural networks,
we are anxious to share these ideas with the broader research community now, instead of waiting for our attempts at achieving validation,
so that the community may participate in the process of discovering its potential and limitations.

\bibliographystyle{apalike}
{\small
\bibliography{refs}}

\end{document}